\documentclass[lettersize,journal]{IEEEtran}
\usepackage{amsmath,amsfonts}
\usepackage{algorithmic}
\usepackage{algorithm}
\usepackage{array}
\usepackage[caption=false,font=normalsize,labelfont=sf,textfont=sf]{subfig}
\usepackage{textcomp}
\usepackage{stfloats}
\usepackage{url}
\usepackage{verbatim}
\usepackage{graphicx}
\usepackage{cite}
\usepackage{color}
\begin{document}
\newcommand{\cui}[1]{{\textcolor{blue}{#1}}}

\title{Style Generation: Image Synthesis based on Coarsely Matched Texts}

\author{Mengyao Cui, Zhe Zhu, Shao-Ping Lu, Yulu Yang}

\maketitle

\begin{abstract}
Previous text-to-image synthesis algorithms typically use explicit textual instructions to generate/manipulate images accurately, but they have difficulty adapting to guidance in the form of coarsely matched texts. In this work, we attempt to stylize an input image using such coarsely matched text as guidance. To tackle this new problem, we introduce a novel task called \textit{text-based style generation} and propose a two-stage generative adversarial network: the first stage generates the overall image style with a sentence feature, and the second stage refines the generated style with a synthetic feature, which is produced by a multi-modality style synthesis module. We re-filter one existing dataset and collect a new dataset for the task. Extensive experiments and ablation studies are conducted to validate our framework. The practical potential of our work is demonstrated by various applications such as text-image alignment and story visualization. Our datasets are published at https://www.kaggle.com/datasets/mengyaocui/style-generation.
\end{abstract}


\section{Introduction}
\label{sec:intro}

Text-to-image synthesis has long been studied. There are two groups of methods for text-to-image synthesis tasks: text-based image generation and text-based image manipulation.
The former usually takes the texts as input and generates a synthesized image as output~\cite{XuZhang-649}.
To impose more controls, layout~\cite{zhao2019image}, scene graphs~\cite{johnson2018image}, and mouse traces~\cite{KohBaldridge-720} are also used as auxiliary input during the image generation process. 
On the contrary, text-based image manipulation only modifies the content of a given image based on the input texts~\cite{DongYu-654,LiQi-647}.
These two categories of methods are implemented with highly matched image-text pairs, where the texts are mostly accurate descriptions without many affective tendencies.
More specifically, the texts of the most commonly used datasets such as COCO~\cite{lin2014microsoft}, Oxford-102 Flowers~\cite{nilsback2008automated}, and CUB-200 Birds~\cite{wah2011caltech} tend to describe the image content directly (e.g. "this bird has a grey crown, a long black bill, and a grey back"). 

\begin{figure}
\includegraphics[width=1\linewidth]{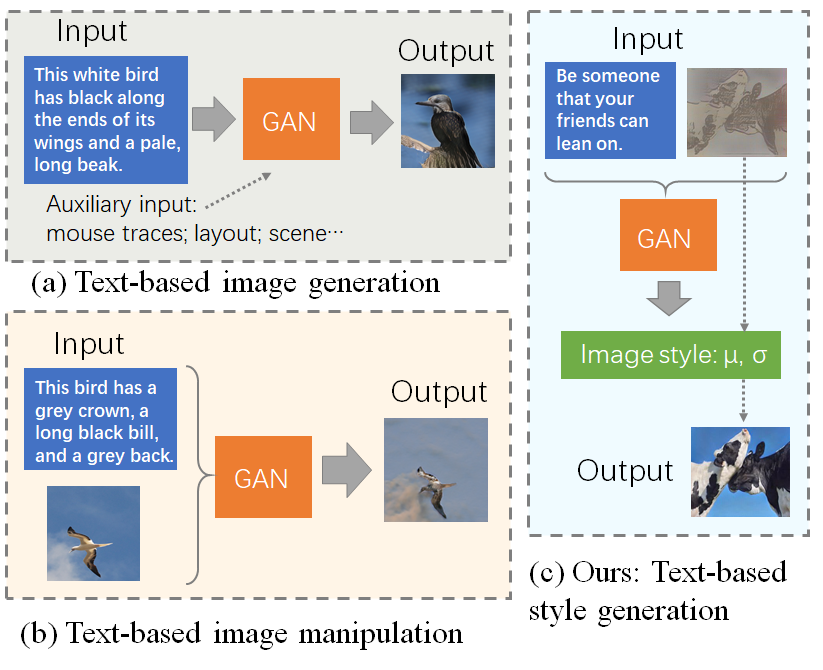}
\caption{Our task vs. traditional text-based image synthesis tasks.}    
\label{fig:diff}
\end{figure}
However, the image-text pairs in real life are not always highly matched. For stories or posts on social media, pictures are often used by authors to convey their feelings, strengthen their standpoint, or supplement the article. In those image-text pairs, despite the fact that the text doesn't describe the image directly, they convey consistent style and emotion. Here we define those pairs as "coarsely matched image-text pairs".
Although coarsely matched image-text pairs are ubiquitous, they are not well studied in existing methods.
For the image generation task in existing literature, the texts are used to generate the image content~\cite{ReedAkata-855,FengNiu-807}, and attention modules are used to precisely match the image and the texts~\cite{XuZhang-649,zhu2019dm}. For the image manipulation task, the texts are used to determine the appearance of the modified area in the image~\cite{LiQi-647,LeeKim-768}, where the correspondence between texts and images is also strong.
Given that in previous methods the text should accurately describe the appearance of the desired output, if the text uses many words that are not accurate descriptions, the image would not be effectively generated (or manipulated). In fact, non-accurate descriptions are common in coarsely matched image-text pairs. Unfortunately,  this type of text is not supported as a guide in image generation and manipulation.

To address the above issue, we introduce a new task called text-based style generation, which aims to generate the style (such as the color and texture) of the content of the image.
We choose to generate the style for coarsely matched texts out of the following considerations. 
Firstly, we only choose to generate the style because instead of synthesizing the whole image, reconstructing the style of the image could overcome the challenge of missing information with vague expressions.
Secondly, the implicit text information, such as an emotion of horror, are strong hints for the style of the synthesized image which can be represented by the overall color and texture of the image~\cite{sutton2016color,chen2020image}.
We show the main difference between our work and previous text-based image synthesis work in Fig.~\ref{fig:diff}. 

\begin{figure*}[t]
\begin{center}
  \includegraphics[width=1\linewidth]{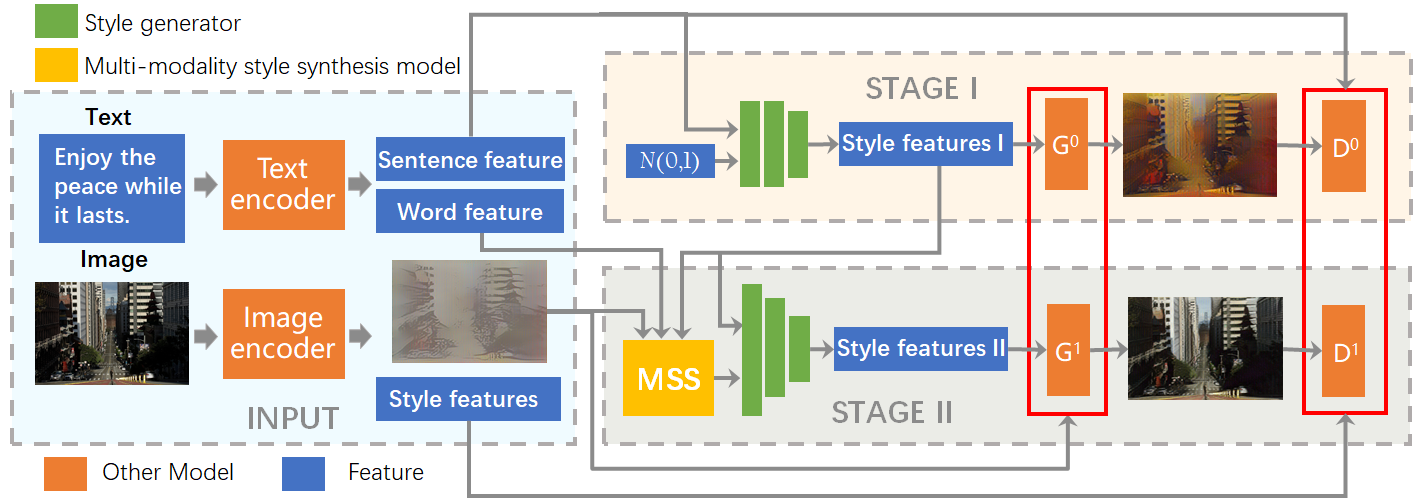}
\end{center}
  \caption{The architecture of our TSG-GAN. In the figure, $MSS$ denotes the multi-modality style synthesis model, $G^{*}$ denotes the image generators and $D^{*}$ denotes the discriminators.}
\label{fig:pipe}
\end{figure*}

In this paper, we propose a novel generative adversarial network for text-based style generation (TSG-GAN), which can generate abundant styles based on the given texts and the existing content of the image. 
To achieve this, we separate the style and content of an image by extracting the content and style features from the image separately, and then we generate a word-level feature and a sentence-level feature for a two-stage process.
In the first stage, the style is generated from the sentence feature, which is concatenated with a random noise. 
In the second stage, we use the word-level feature, the image content feature, and the first-stage style features as the input of a multi-modality style synthesis module. The module combines the word-level feature and the first-stage image feature with the attention mechanism and generates the synthesis feature. The synthesis feature is then concatenated with the first-stage style features and later used for the second-stage generation.
To validate our method, we establish coarsely matched image-text datasets by re-filter the Contextual Caption dataset~\cite{ChowdhuryBhowmik-758} and collecting a new dataset named coarse text-illustration dataset.
We apply extensive ablation studies to verify the necessity of each module of our framework.
The qualitative and quantitative comparisons show that our framework performs better than state-of-the-art text-based image synthesis methods on coarsely matched texts. 

Our work can be widely used on image synthesis tasks with such coarsely matched texts.
We show some examples of text-image alignment~\cite{chowdhury2021sandi} and story visualization~\cite{LiGan-661,Song-663}.

In summary, our contributions are listed as follows.

\begin{itemize}
\item We propose a novel style generation task with coarsely matched texts.  
\item We provide a novel generative adversarial network for text-based style generation, which consists of two generating stages based on the attention mechanism. 

\item We establish datasets for the style generation task and illustrate the potential of our work on text-image alignment and story visualization.
\end{itemize}

\section{Related work}
\label{sec:relate}
\paragraph{Text-based image generation.} 
Text-based image generation produces images based on the input text.
In~\cite{ReedAkata-855}, The conditional GAN is firstly introduced to implement text-guided image generation, where the whole sentence embedding is conditioned from given texts. 
Stacked generators and discriminators are then used to synthesize images with higher resolution and quality~\cite{ZhangXu-851}.
To better focus on different image regions corresponding to the texts, the attention mechanism is introduced into the generation model~\cite{XuZhang-649}, and some new frameworks with dynamic memory networks are also proposed~\cite{zhu2019dm}. 
Later, an attentional self-modulation generator is further designed to maximize the mutual information between image and text~\cite{ZhangKoh-806}. 
Recently, a method~\cite{FengNiu-807} disentangles the content and the style during generation, where the content of the image is generated from texts, and the style of the image is generated from a random noise or a reference image.

Additional supervision can also benefit image generation. 
To process texts with multi-objects and complex relationships, a scene graph is used for image generation with a graph convolution network~\cite{johnson2018image}. 
Spatial layout, consisting of bounding boxes and object categories is also used for generating images with multi-objects~\cite{zhao2019image}. 
To implement interactive image generation, TRECS~\cite{KohBaldridge-720} exploits mouse traces as an auxiliary
input, which first generates and composes scene masks, and then translates masks to images.

\paragraph{Text-based image manipulation.} 
Different from text-based image generation, text-based image manipulation takes the original image as an extra input, trying to preserve the text-irrelevant part of the image.
The conditional GAN framework is pioneeringly used by \cite{DongYu-654} on text-based image manipulation. 
Then the word-level information is considered with a text-adaptive discriminator~\cite{NamKim-655}. 
In \cite{LiQi-647}, the word-level information is further utilized to select image regions with a text-image affine combination module and generate a photo-realistic result with a detail correction module. 
Styleclip~\cite{PatashnikWu-686} uses the Contrastive Language-Image Pre-training models~\cite{radford2021learning} to match the image and text, then provides three methods for generating realistic images based on the high performance of StyleGAN~\cite{KarrasLaine-583}. 
Similar to text-based image generation, an auxiliary input is also used in text-based image manipulation. 
Take \cite{dhamo2020semantic} as an example, a semantic scene graph is first predicted from the source image, and then used for interactive image manipulation. 

Different from these tasks, our work generates the style of the image based on coarsely matched text information.
To clarify, although the Adain module~\cite{huang2017arbitrary} is used in many image manipulation tasks~\cite{KarrasLaine-583, PatashnikWu-686, liu2020describe}, they do not actually view the image as a combination of the style and the content. 
For instance, in \cite{KarrasLaine-583}, the "style" is calculated from a vector (instead of an example image) and is manipulated under multiple resolutions.

\paragraph{The content and the style of images.}
Tasks such as image style transfer~\cite{huang2017arbitrary, li2017universal, JingLiu-726}, image domain adaptation~\cite{yang2020fda}, and generalization~\cite{xu2021fourier} take images as a combination of the style and the content. 
There are several popular ways to separate the style and the content of the image.
In Adaptive Instance Normalization~\cite{huang2017arbitrary,JingLiu-726}, the style is viewed as the channel-wise mean and variance of the image, which are directly calculated from the input feature.
Later, dynamic instance normalization is introduced to provide more flexible transfers~\cite{jing2020dynamic}.
Another way is to apply the whitening transform to remove the style from the image and use the coloring transform to generate the new style~\cite{li2017universal}. 
The third way, proposed by Avatarnet~\cite{sheng2018avatar}, implements style transfer by matching patches in the style image and the content image.
Last but not least, some other methods encode the image feature to a latent code that concatenates the content and the style feature~\cite{svoboda2020two,FengNiu-807}. 
Recently, a lossless and reversible network is proposed to solve the serious content leak problem suffered by earlier methods, which achieves unbiased style transfer~\cite{AnHuang-800} based on neural flow.

\section{GAN for text-based style generation}
\label{sec:appro}
\subsection{Architecture}
Let $v\in \mathbb{R}^{C\times H\times W}$ denote the content feature of a given image, and $\mu, \sigma\in \mathbb{R}^C$ denote image style features, where $C$ is the channel number, $H$ and $W$ are the height and width of the content feature, respectively.  
Let $t$ be coarsely matched texts.
The aim of our work is to generate the target style $\mu, \sigma$ based on text-content pairs ($t$, $v$).
We first use the sentence-level feature $\bar{e}\in \mathbb{R}^{D}$, which provides overall information of $t$, to generate preliminary styles $\mu_0, \sigma_0$.  
Observing that both the content feature $v$ and the text feature $t$ have an effect on the style of the image, we use a multi-modality style synthesis model in the second generation stage. Specifically, we design a cross-attention block and a self-attention block. The blocks take the first-stage image feature $v_0\in \mathbb{R}^{C\times H\times W}$ and the word-level feature $e\in\mathbb{R}^{D\times T}$ of $t$ as input.
The output feature $o\in\mathbb{R}^{2*C}$ is then used to generate fine-grained styles $\mu_1, \sigma_1$. 
The framework of our TSG-GAN is shown in Fig~\ref{fig:pipe}.

\begin{figure*}[t]
\begin{center}
  \includegraphics[width=1\linewidth]{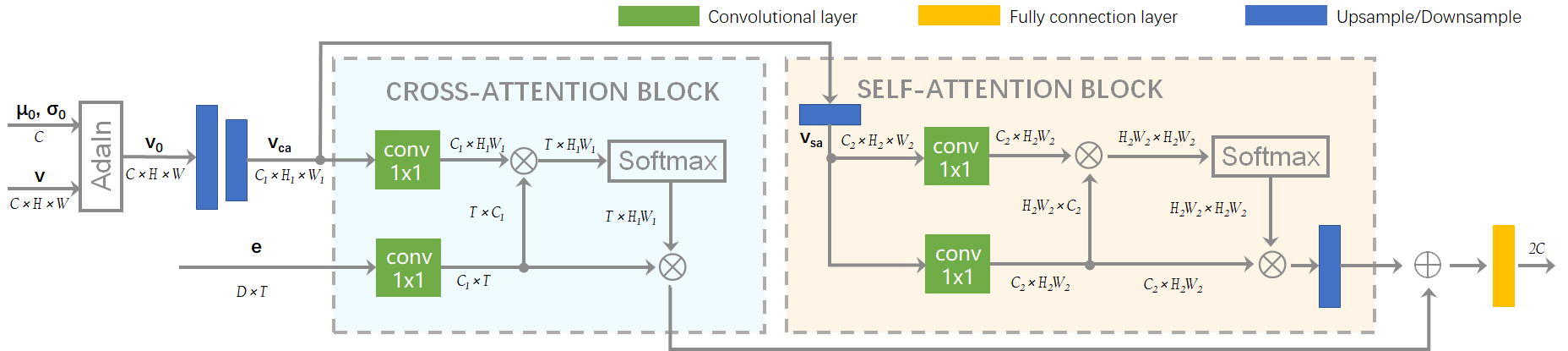}
\end{center}
  \caption{The proposed multi-modality style synthesis model. The model consists of a cross-attention block and a self-attention block. Before being fed into two blocks, the image style features $\mu_0,\sigma_0$ and the image content feature $v$ are first combined with the Adain module, then processed with several down-sampling blocks.}
\label{fig:attn}
\end{figure*}

\subsubsection{Style generation model}
Our style generation model contains two style generators ($SG_0$ for stage 1, $SG_1$ for stage 2) to generate the style feature, and one multi-modality style synthesis model ($F^{MSS}$ for stage 2). 
In the first stage, sentence embedding $\bar{e}$ is transferred to the conditioning vector $\bar{e_c}\in \mathbb{R}^{c}$ by the conditioning augmentation~\cite{ZhangXu-851}, and a noise vector $z$ is sampled with a standard normal distribution.
Specifically,
\begin{equation}
\begin{aligned}&\mu_0, \sigma_0 = SG_0(z,\bar{e_c}), \\
&\mu_1, \sigma_1 = SG_1(\mu_0, \sigma_0 , F^{MSS}(e,v,\mu_0, \sigma_0)).
\end{aligned}
\end{equation}
In $SGs$, the input vectors are first concatenated, and then fed into a network that consists of three fully connected layers to generate style features.
As shown in Fig.~\ref{fig:attn}, the $F^{MSS}$ contains a cross-attention block and a self-attention block. First, we combine the content feature $v$ and the style features $\mu_0,\sigma_0$ with \cite{huang2017arbitrary}, and get the new image feature $v_0\in \mathbb{R}^{C\times H\times W}$. We reduce the size of $v_0$ by down-sampling and extracting higher-level feature $v_{ca} \in \mathbb{R}^{C_1\times H_1\times W_1}, v_{sa} \in \mathbb{R}^{C_2\times H_2\times W_2}$. 
In existing text-based image synthesis work~\cite{XuZhang-649}, only a cross-attention block is used, which provides a word-context image feature. We make an improvement by using an additional self-attention block to provide more information for overall style generation.

More specifically, we generate synthesis style feature $o\in \mathbb{R}^{2C}$ with a cross-attention block and a self-attention block as follows:

\begin{equation}
\small
o = F^{MSS}(e,v,\mu_0, \sigma_0) = FC(\phi_c(v_{ca},e) + \phi_s(v_{sa})),
\end{equation}
where $\phi_c$ denotes cross-attention block, $\phi_s$ denotes an image self-attention block, and $FC$ represents a fully connected layer.

In attention blocks, the image feature $v_{ca}, v_{sa}$ and the word feature $e$ are first weighted by $1\times 1$ convolution layers:

\begin{equation}
\small
\begin{aligned}&v_c=\chi_0(v_{ca}), v_c \in \mathbb{R}^{C_1\times H_1\times W_1};\\
&e_c=\chi_1(e), e_c \in \mathbb{R}^{C_1\times T};\\
&v_q=\chi_2(v_{sa}), v_q \in \mathbb{R}^{C_2\times H_2\times W_2};\\
&v_k=\chi_3(v_{sa}), v_k \in \mathbb{R}^{C_2\times H_2\times W_2}.
\end{aligned}
\end{equation}
In the above equations, $\chi_*(\cdot)$ denote $1\times1$ convolutional layers.

\begin{figure}
\includegraphics[width=1\linewidth]{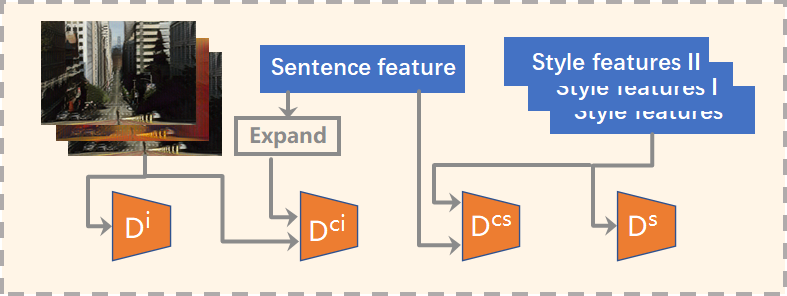}
\caption{The proposed discriminator consists of four branches: $D^i,D^{ci},D^s,D^{ci}$, which take the image feature, the sentence feature, and the style features as input.}   
\label{fig:dis}
\end{figure}

After that, we compute the similarity matrices and normalize them with Softmax~\cite{XuZhang-649}.
\begin{equation}
\small
\begin{aligned}&C=Softmax(e_{c}^Tv_{c}), C \in \mathbb{R}^{T\times H_1W_1},\\
& S=Softmax(v_{q}^Tv_{k}), S \in \mathbb{R}^{H_2W_2\times H_2W_2},
\end{aligned}
\end{equation}
where $C$ is the similarity for all pairs of words in the sentence and sub-regions in the image, and $S$ is the similarity for all pairs of sub-regions in the image. Both of the metrics are calculated by the dot product.

Then we calculate the word-context image vector and the image self-attention vector with the following equations:

\begin{equation}
\begin{aligned}&\phi_c(v_{ca},e) = e_cC,\\
&\phi_s(v_{sa})) = \epsilon(v_qS).
\end{aligned}
\end{equation}
In this equation, $\epsilon(\cdot)$ up-sample the feature $v_qS$ to the same size as $\phi_c(v_{ca},e)$.

Similar to $v_0$, we generate $v_1\in \mathbb{R}^{D\times H \times W}$ with $\mu_1, \sigma_1$ using the Adain module. With the reverse inference of Artflow, $v_0, v_1$ are then transferred to the images $I_0, I_1$. 

\subsubsection{The discriminator network} 
Our discriminator network is designed to have four branches: $D^s, D^i, D^{cs}, D^{ci}$ (see Fig.~\ref{fig:dis}). 
In the network, $D_s$ judges the visual reality of the image style, $D_i$ judges the visual reality of the image itself, $D_{cs}$ evaluates the semantic consistency of style-text pairs, and $D_{ci}$ evaluates the semantic consistency of image-text pairs. 
Their corresponding output scores $s_s, s_i, s_{cs}, s_{ci}$ are calculated as follows:
\begin{equation}
\begin{aligned}
& s_i(\hat{x})=D^i(I_i), s_i(x)=D^i(I),\\
& s_s(\hat{x})=D^s(\mu_i,\sigma_i),  s_s(x)=D^s(\mu,\sigma), \\
& s_{ci}(\hat{x})=D^{ci}(I_i, \bar{e}), s_{ci}(x)=D^{ci}(I, \bar{e}),\\
& s_{cs}(\hat{x})=D^{cs}(\mu_i,\sigma_i, \bar{e}),  s_{cs}(x)=D^{cs}(\mu,\sigma,\bar{e}),
\end{aligned}  
\end{equation}
where $i$ means $i^{th}$ generation of the work, $x$ represents the real image distribution, $\hat{x}$ denotes the model distribution. 
\subsection{Objective function} 
Our objective function consists of two types of losses: the adversarial loss $L_G$ and $L_D$, and the style loss $L_s$.

\subsubsection{The adversarial loss} We employ two conditional losses (based on $s_{cs}$, $s_{ci}$) and two unconditional losses (based on $s_i, s_s$) to train the model, which are listed as follows:
\begin{equation}
\small
\begin{aligned}
L_G = &-\mathbb{E}_{\hat{x}\sim p_G}[log(s_s(\hat{x}))]  -\mathbb{E}_{\hat{x}\sim p_G}[log(s_i(\hat{x}))]\\
&-\mathbb{E}_{\hat{x}\sim p_G}[log(s_{cs}(\hat{x}))]  -\mathbb{E}_{\hat{x}\sim p_G}[log(s_{ci}(\hat{x}))],\\
\end{aligned}  
\end{equation}
\begin{equation}
\small
\begin{aligned}
L_D = &-\mathbb{E}_{x\sim p_{data}}[log(s_s(x))]  -\mathbb{E}_{\hat{x}\sim p_G}[log(1-s_s(\hat{x}))]\\
&-\mathbb{E}_{x\sim p_{data}}[log(s_i(x))]  -\mathbb{E}_{\hat{x}\sim p_G}[log(1-s_i(\hat{x}))]\\
&-\mathbb{E}_{x\sim p_{data}}[log(s_{cs}(x))]  -\mathbb{E}_{\hat{x}\sim p_G}[log(1-s_{cs}(\hat{x}))]\\
&-\mathbb{E}_{x\sim p_{data}}[log(s_{ci}(x))]  -\mathbb{E}_{\hat{x}\sim p_G}[log(1-s_{ci}(\hat{x}))],
\end{aligned}  
\end{equation}
where $x$ represents the real image distribution $p_{data}$, and $\hat{x}$ represents the model distribution of $p_G$.

\begin{figure*}[t]
\begin{center}
  \includegraphics[width=1\linewidth]{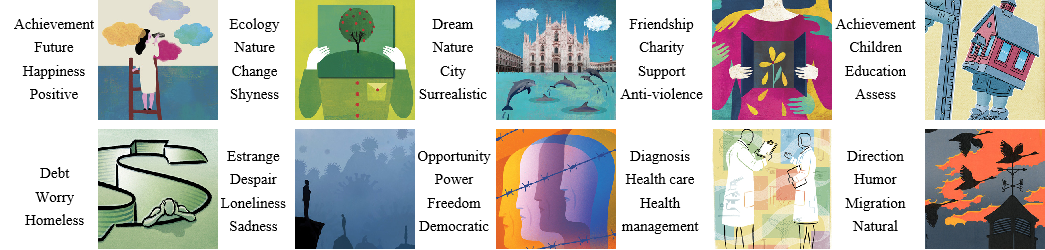}
\end{center}
   \caption{Examples of our coarse text-illustration dataset. The dataset consists of 10317 images, each image corresponding to several coarsely matched texts.}
\label{fig:dataset}
\end{figure*}

\begin{figure*}[t]
\begin{center}
  \includegraphics[width=1\linewidth]{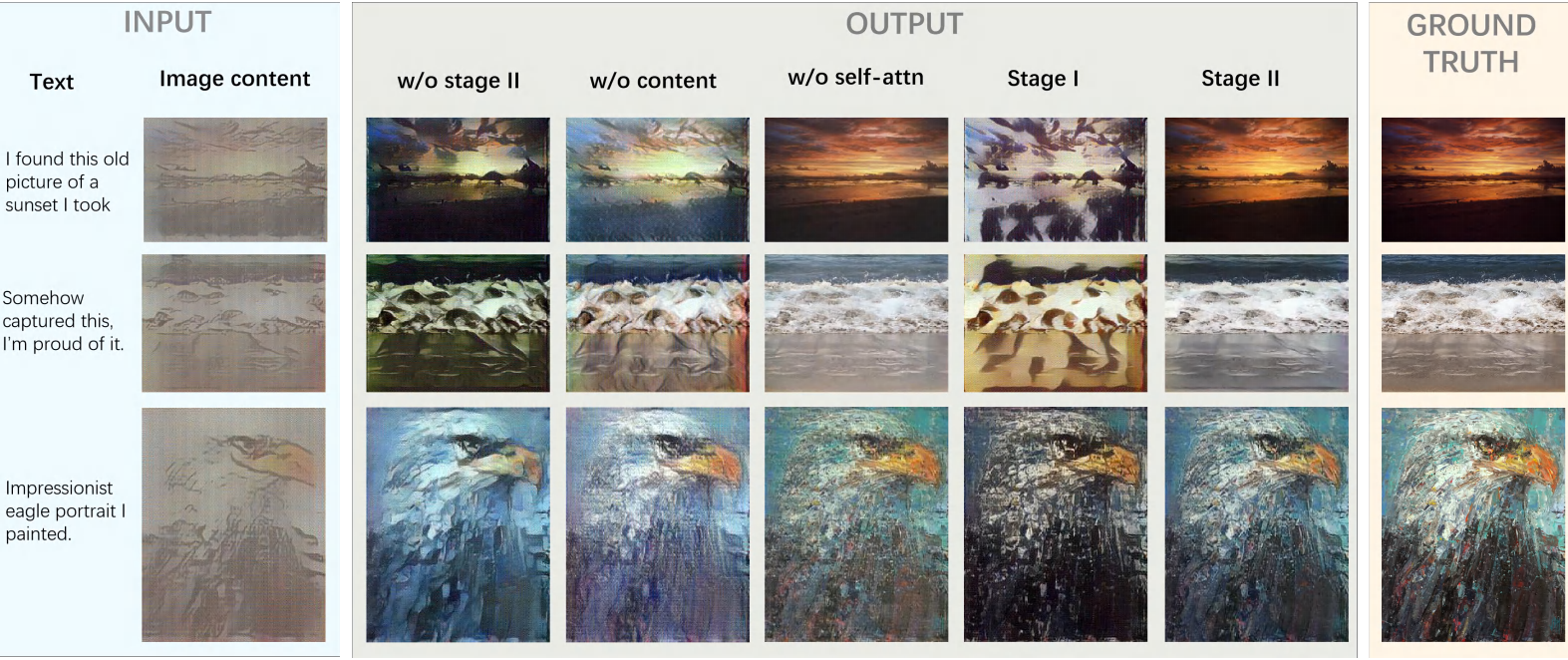}
\end{center}
   \caption{Ablation studies on the Contextual Caption dataset. "w/o self-attn" represents only using cross-attention block in the second generating stage. "w/o content" denotes training without $v_c$ in the second generating stage. For "w/o stage II", we totally remove the second generating stage. "stage I" shows the generating result in the first stage of the full model, while "stage II" shows the generating result in the second stage of the full model.}
\label{fig:ablation}
\end{figure*}

\begin{figure*}[t]
\begin{center}
   \includegraphics[width=1\linewidth]{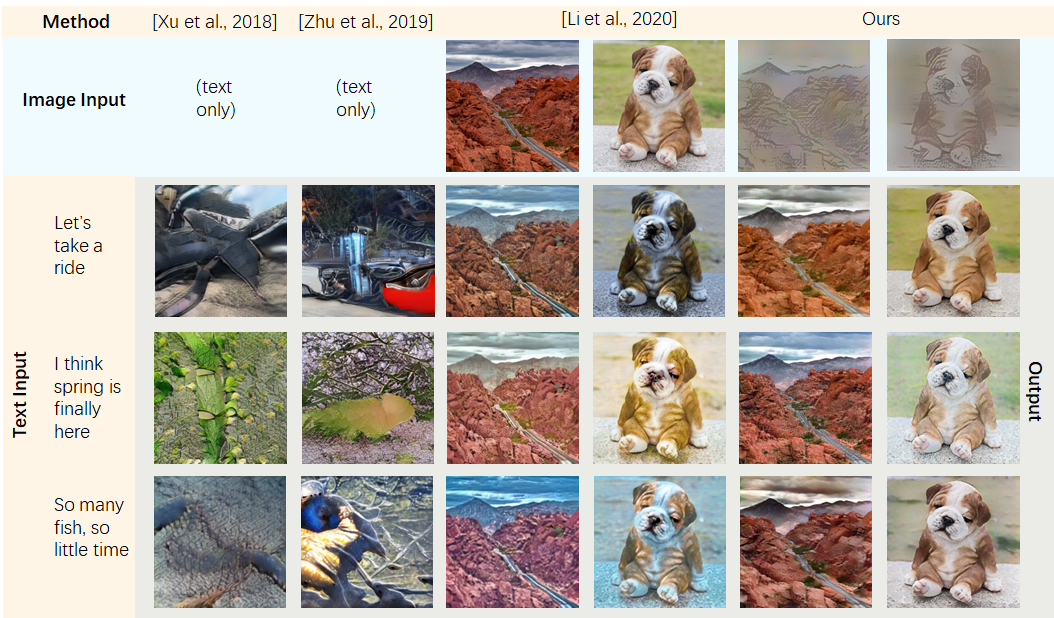}
\end{center}
   \caption{Qualitative comparison of four methods on the Contextual Caption dataset. The first column shows the result of \cite{XuZhang-649}, and the second column shows the result of \cite{FengNiu-807}. The third and fourth columns show the result of \cite{LiQi-647}, and the last two columns show the result of ours.}
\label{fig:compare}
\end{figure*}

\subsubsection{The style loss}
Following \cite{FengNiu-807}, we measure the style loss between the ground truth style and the generated style using Pearson correlation $\rho$:
\begin{equation}
L_S = -\rho(h, h_i),
\end{equation}
where $h$ is the concatenated $\mu$ and $\sigma$, $h_i$ is the concatenated $\mu_i$ and $\sigma_i$. Finally, we get the total loss of our TSG-GAN:
\begin{equation}
\begin{aligned}
&L_{D_{total}} = L_D + \lambda L_s,\\
&L_{G_{total}} = L_G + \lambda L_s,
\end{aligned}
\end{equation}
where $\lambda$ is a hyper-parameter to balance the terms of the loss functions.

\section{Experiments}
\label{sec:expe}
We carry out extensive experiments to evaluate the proposed TSG-GAN. 
First, we design an ablation study to evaluate the main components of our framework: the attention module, the style generation module, and the discriminator.  
Then we compare our model with several other state-of-art methods~\cite{XuZhang-649,FengNiu-807,LiQi-647}.
All the above-mentioned models are trained and tested on the Contextual Caption dataset. 

\paragraph{Datasets.} Popular image-text datasets are constructed from the news reports, which contain formal expressions with no affective description~\cite{lin2014microsoft,nilsback2008automated,wah2011caltech}. 
For our style generation task, instead of using previous widely-used datasets, we re-filter one existing dataset and collect a new dataset.
Specifically, we employ the Contextual Caption dataset~\cite{ChowdhuryBhowmik-758}, which provides more informal and affective descriptions for images. 
Since the dataset is directly crawled from social media, we re-select the dataset to make it more suitable for our task.
Firstly, we remove the images with broken links or invalid captions (captions with less than five valid characters) from the original Contextual Caption dataset.
Then we manually remove images containing insufficient style information, such as monochromatic images and text-dominant images.
We also remove images containing inappropriate content such as pornography, blood, and other potentially harmful guiding content.
After removal, we get 69,284 remaining images in total. 
We randomly divide four-fifths of the dataset into the
training set and one-fifth of the dataset into the
validating set, where each image is associated with one text description. 

We build a dataset collected from a coarse text-illustration dataset website~\footnote{\url{https://www.theispot.com/}}, where each illustration is described by its author with several keywords.
Our coarse text-illustration dataset contains $10317$ images in total, we show some samples of the dataset in Fig.~\ref{fig:dataset}.  
Similarly, we randomly divide four-fifths of the dataset into the
training set and one-fifth of the dataset into the
validating set. 
To prevent overfitting, the images are randomly cropped and flipped, then resized to $(256, 256)$.

\paragraph{Implementation.} 
The parameter settings of our text encoder are the same as \cite{XuZhang-649}. 
We use LSTM to extract text embedding, including a word-level embedding $e\in\mathbb{R}^{256\times 18}$, and a global sentence embedding $\bar{e}\in \mathbb{R}^{256}$. 
The length of the conditioning vector $e_c$ and noise $z$ are set to 100.
To implement style generation in an unbiased way, we use a reversible encoder Artflow~\cite{AnHuang-800} as our image encoder.
In detail, we use the pre-trained Artflow module to encode the given image $I\in \mathbb{R}^{3\times 256\times 256}$.
Then we extract the content feature $v\in \mathbb{R}^{48\times 32\times 32}$, and the style features $\mu, \sigma\in \mathbb{R}^{48}$ from the image $I$ with the Adain module.
The Artflow module is set to evaluation mode during the whole training process. 
After the separation, $(c, T)\rightarrow s$ pairs are used to train our model and construct the loss function. 
Our method has two generation stages, each containing a generator and a set of discriminators. 
In the multi-modality style synthesis model, we first downscale the spatial size of the image and get
$v_{ca}\in \mathbb{R}^{128\times 16\times 16}, v_{sa}\in \mathbb{R}^{256\times 8\times 8}$. Then the synthesis feature $o\in \mathbb{R}^{96}$ is generated by two attention blocks.
We employ a similar discriminator architecture as \cite{XuZhang-649}, where the kernel size and the stride of the convolution layer are changed according to the input feature size. 
In $D^s$, $\mu_i$ and $\sigma_i$ are concatenated before being fed into the convolution layer. 
In $D^{cs}$, we concatenate $\mu_i$, $\sigma_i$, and the global sentence embedding $\bar{e}\in \mathbb{R}^{256}$, then feed them into the joining convolution layer.
The whole model is trained for 160 epochs on the Contextual Caption dataset~\cite{ChowdhuryBhowmik-758}. 
We use the Adam optimizer~\cite{kingma2014adam} with $\beta_1 = 0.5$ and $\beta_2 = 0.999$. 
The learning rate of the generators, the discriminators, and the text encoder are 0.0002, 0.0002, and 0.002, respectively. 
The hyper-parameter $\lambda$ is set to 0.1.

\subsection{Ablation study}
\paragraph{Evaluating metrics.}
We evaluate our TSG-GAN using the precision of the generated style, the image quality, and the diversity. 
For the image quality and diversity, we choose the inception scores ($IS$) and Fr$\acute{e}$chet Inception Distance ($FID$) to evaluate them. 
Peak Signal to Noise Ratio ($PSNR$) is used to further evaluate the image quality. 
Besides, the coherence of the generated style and the text also needs to be measured, thus we design a style loss metric.
For the target style $s$ and the generated style $s'$, we calculate the style loss ($SL$) as follows:
\begin{equation}
SL = \frac{1}{C}||s'-s||_2,
\end{equation}
where $C$ is the channel number of $s, s'$. 

\begin{table}
\centering
\caption{Ablation study: inception scores ($IS$), Fr$\acute{e}$chet Inception Distance ($FID$), Style Loss ($SL$), Peak Signal to Noise Ratio ($PSNR$) of our approaches on the Contextual Caption dataset. Among the indexes, higher $IS$ and $PSNR$, and lower $SL$ and $FID$ mean better results.
}
\begin{tabular}{lrrrrlrrrr}
\hline
Method & IS & FID & SL & PSNR\\
\hline
w/o stage II & 10.6 & 47.9 & 0.0357 &59.8 \\
w/o content & 10.1 & 47.1 & 0.0442 & 61.5  \\
w/o self-attn & 18.9 & 9.96 & 0.0181 & 67.3 \\
w/ $D^s, D^{cs}$ & 16.3 & 16.9 & 0.0171 & 64.6 \\
full model & \textbf{19.8} & \textbf{4.87} & \textbf{0.0163} & \textbf{68.9} \\
\hline
\end{tabular}
\label{tab:ablation}
\end{table}

\paragraph{Ablation studies for the two-stage style generation.}
We show the generated images in two generating stages separately (see Fig.~\ref{fig:ablation}). 
In the first stage, preliminary contours of colors and textures are generated, which are further optimized in the second stage.
We also provide the results with only a single generating stage in the second column of Fig.~\ref{fig:ablation}, to further verify the effectiveness of the two-stage generation. 
Here the images lose most of the details and are difficult to distinguish, the evaluating metrics are also significantly worse than those with the full model (see Tab.~\ref{tab:ablation}). 

\paragraph{Effectiveness of the proposed multi-modality style synthesis model.} 
Different from traditional attention solutions, our multi-modality style synthesis model uses an additional self-attention block.  
To validate the effectiveness of this block, we train another model without a self-attention block. As shown in Fig.~\ref{fig:ablation}, in the third column of the output, the texture of "w/o self-attn" is worse than the full model.
The evaluating scores also get worse (see Tab.~\ref{tab:ablation}).

To further illustrate the effectiveness of the image content feature $v$, we remove the multi-modality style synthesis model and just use the word feature $e$ for the second-stage generation. 
As we can see, in the second column of the output in Fig.~\ref{fig:ablation}, the generated images lose great details and look unrealistic.
Another evidence of the effectiveness of using the image content feature can be found in Tab.~\ref{tab:ablation}.
More analysis is shown in supplementary materials.

\paragraph{Effectiveness of the discriminator.} Traditional text-image generation methods (eg.\cite{XuZhang-649}) take two branches $D^i, D^{ci}$ in discriminator.
For style generation, we use $D^s, D^{cs}$ as basic branches of the discriminator, and $D^i, D^{ci}$ are further added to improve the visual quality of the generated image. 
As shown in Tab.~\ref{tab:ablation}, the quality of generated images significantly declines when missing $D^i$ and $D^{ci}$. 
We show more analysis of discriminators in supplementary materials.

\subsection{Comparison with state-of-art methods}
\subsubsection{Qualitative comparison}
As stated before, previous image-text synthesis methods are designed to generate or manipulate the image from explicit text descriptions. 
Now we compare those methods with our work on the Contextual Caption dataset.
Specifically, we select two methods for text-based image generation~\cite{XuZhang-649,FengNiu-807}, and another method for text-based image manipulation~\cite{LiQi-647}.
As shown in Fig.~\ref{fig:compare}, image generation methods~\cite{XuZhang-649,FengNiu-807} only take the texts as input, while the image manipulation method~\cite{LiQi-647} takes the texts and the original image as input. 
Different from them, our work takes the texts and the content of the image as input. 
Image generation methods~\cite{XuZhang-649,FengNiu-807} can barely generate meaningful pictures when there are no clear text instructions (see Fig.~\ref{fig:compare}, columns 1 and 2 of the output). 
Although the method of \cite{LiQi-647} is capable of generating higher quality images than image generation methods, it fails to preserve or generate convincing textures.
One can see that in Fig.~\ref{fig:compare}, column 3, row 2 of the output, the texture and color of the mountain is distorted.
Additionally, the generated style of the dog is neither pleasant nor well-aligned with the text in column 4.

On the contrary, our method can generate images with high visual quality, and it is also capable of generating new colors and textures. 
To sum up, our method is more suitable for the image synthesis task with coarsely matched texts.
More examples can be found in supplementary materials.

\subsubsection{Quantitative comparison}

It is difficult to complete a fair comparison between style generation, image generation, and image manipulation work with existing metrics.
Specifically, since our method has different input from image generation and image manipulation work, it is unfair to use inception scores ($IS$) or Fr$\acute{e}$chet Inception Distance ($FID$) directly. 
Peak Signal to Noise Ratio ($PSNR$) is also not suitable, since there is no ground truth for most image synthesis methods.
Style loss ($SL$) is also not a valid index for image synthesis methods. 
Thus we design a user study to quantitatively demonstrate our work can reach higher performance on coarsely matched texts.

\paragraph{Procedure and materials}
We design two types of user studies here.
User study A is used to measure the degree of the correspondence between the generated images and their original texts.
Specifically,  we select 32 image-text pairs from the evaluation part of the dataset. 
For each method, we generate 32 images based on the 32 inputs. 
Here we get 32 text-image pairs for each method, where one text corresponds to 4 images generated by 4 methods.
In the questionnaire, the participants show their preference for the generated images by answering the following question:

\begin{itemize}
\item Please select the picture which matches the text best.  
\end{itemize}

In user study B, we further examine if our method can generate consistent styles with different input texts. 
In detail, we select 8 images and 8 texts separately from the evaluation part of the dataset.
We match each text with 8 images separately and get 64 image-text pairs.
Taking those image-text pairs as input, we then generate 64 images with our method and Manigan~\cite{LiQi-647} separately.
Each text corresponds to 8 images for each method. 
In the questionnaire, each participant is given 8 texts, and each text is shown with two of their corresponding images (randomly selected from both of the methods).
We evaluate the coherence of the image-text pairs and the preference of participants with the following questions:

\begin{itemize}
\item Please evaluate the consistency between the style of image A and the text. (0 very inconsistent, 4 very consistent);
\item Please evaluate the consistency between the style of image B and the text. (0 very inconsistent, 4 very consistent);
\item Please select the picture which matches the text best.  
\end{itemize}

In questions $1$ and $2$, a 5-point Likert scale is used. The materials of the user study can be found in supplementary materials.

\begin{figure}[t]
\begin{center}
 \includegraphics[width=1\linewidth]{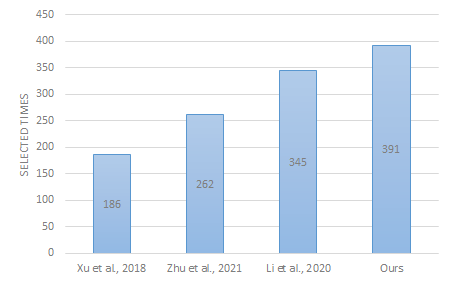}
\end{center}
   \caption{The selected times in user study A.}
\label{fig:ave}
\end{figure}

\begin{figure}[t]
\begin{center}
 \includegraphics[width=1\linewidth]{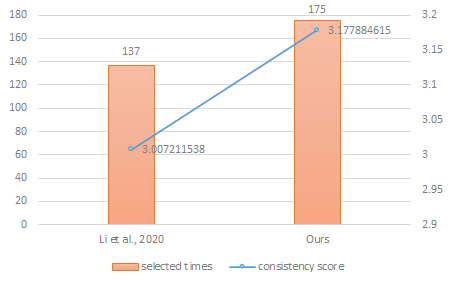}
\end{center}
   \caption{The selected times and consistency scores in user study B.}
\label{fig:ave2}
\end{figure}

\paragraph{Participants and results}
We published our questionnaires. 
In user study A, we received 47 answers, among which 37 answers were valid.
As shown in Fig.~\ref{fig:ave}, the images of our method have been chosen 391 times (33.02\%), and the images of \cite{LiQi-647} have been chosen 345 times (29.12\%). 
Similarly, the participants chose the images of \cite{FengNiu-807} 262 times (22.12\%) and \cite{XuZhang-649} 186 times (15.71\%).
We use the Chi-square test to test the significance of the difference. 
The result shows the difference between our method and  \cite{LiQi-647,FengNiu-807,XuZhang-649} is significant (p<0.01). 
In all, the participants significantly prefer the generated images from our method to match the texts.

The questionnaire of user study B received 53 answers, among which 52 answers were valid.
As shown in Fig.~\ref{fig:ave2}, the images of our method have been chosen 175 times (56.09\%), and the images of \cite{LiQi-647} have been chosen 137 times (43.91\%). 
We also use the Chi-square test here to test the difference between the two methods at selected times.
It turns out the difference between our method and \cite{LiQi-647} is significant (p<0.01).
Besides, the consistency score of \cite{LiQi-647} is $3.01$, and the consistency score of our method is $3.18$ (see Fig.~\ref{fig:ave2}). 
The consistency score of samples doesn't follow normal distribution according to the Kolmogorov–Smirnov test. 
Thus we use the Kruskal-Wallis H test to compare our method with \cite{LiQi-647}.
The score of our method is significantly higher than \cite{LiQi-647} (p<0.1), which means our method is thought to generate more consistent images with the input text.

To summarize, the result of the user study shows that our method can generate more consistent styles of images with the given text.

\begin{figure}[t]
\begin{center}
 \includegraphics[width=1\linewidth]{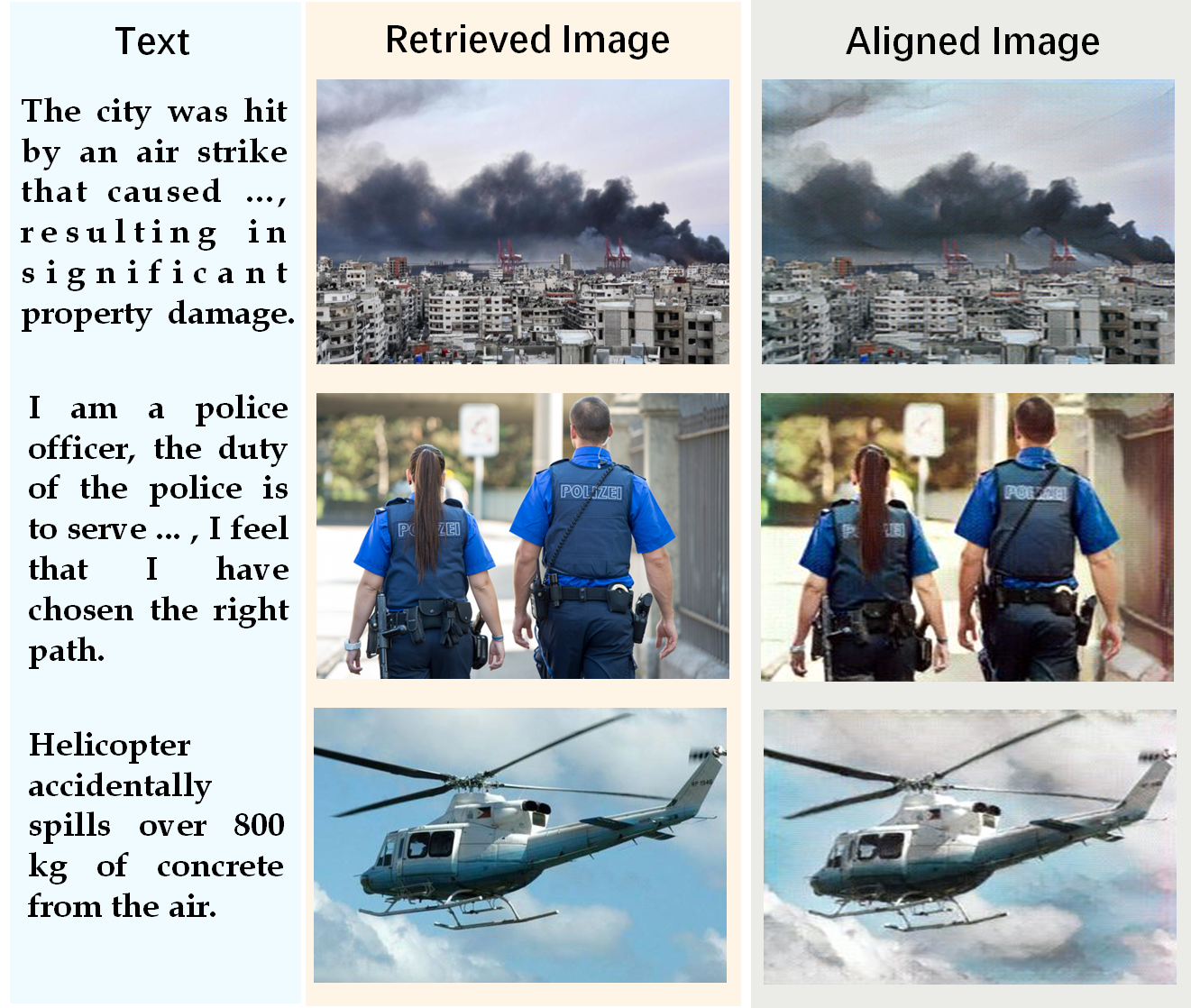}
\end{center}
   \caption{Examples of text-image alignment.}
\label{fig:align}
\end{figure}

\section{Applications}
\label{sec:appl}
Here we show some potential applications of our work on text-image alignment and story visualization tasks.

\subsection{Text-image alignment}
Given a document, the text-image alignment task first selects images from a dataset, then decides where the images are placed~\cite{LiuLebret-679,chowdhury2021sandi,ZhangWang-746}.
Based on the document type, text-image alignment work can be divided into story-image alignment~\cite{chowdhury2021sandi}, where images are more used for supplementary narration, and news-image alignment~\cite{LiuLebret-679,ZhangWang-746}, where images are more used for general illustration.
However, the retrieved images may not be well aligned with the texts on affection or other implied information. 

Our work can be used here to improve the alignment by modifying the overall styles. 
Fig~\ref{fig:align} shows some examples of the application for text-image alignment, where the original image is retrieved by an online API~\footnote{\url{https://modemos.epfl.ch/article}}.
To be specific, we respectively show images with negatively (rows 1, 3) and positively (row 2) impacted texts, and the aligned images seem more consistent with the texts than the retrieved images.

\subsection{Story visualization}

Existing story visualization methods~\cite{LiGan-661,Song-663} usually aim at generating coherent cartoon characters and their corresponding scenes for stories. 
We provide another way to implement story visualization. 
First, users select (or retrieve) the target image based on the input texts. 
Secondly, our method generates the style of the story based on the content of the image.  

\begin{figure}[t]
\begin{center}
   \includegraphics[width=1\linewidth]{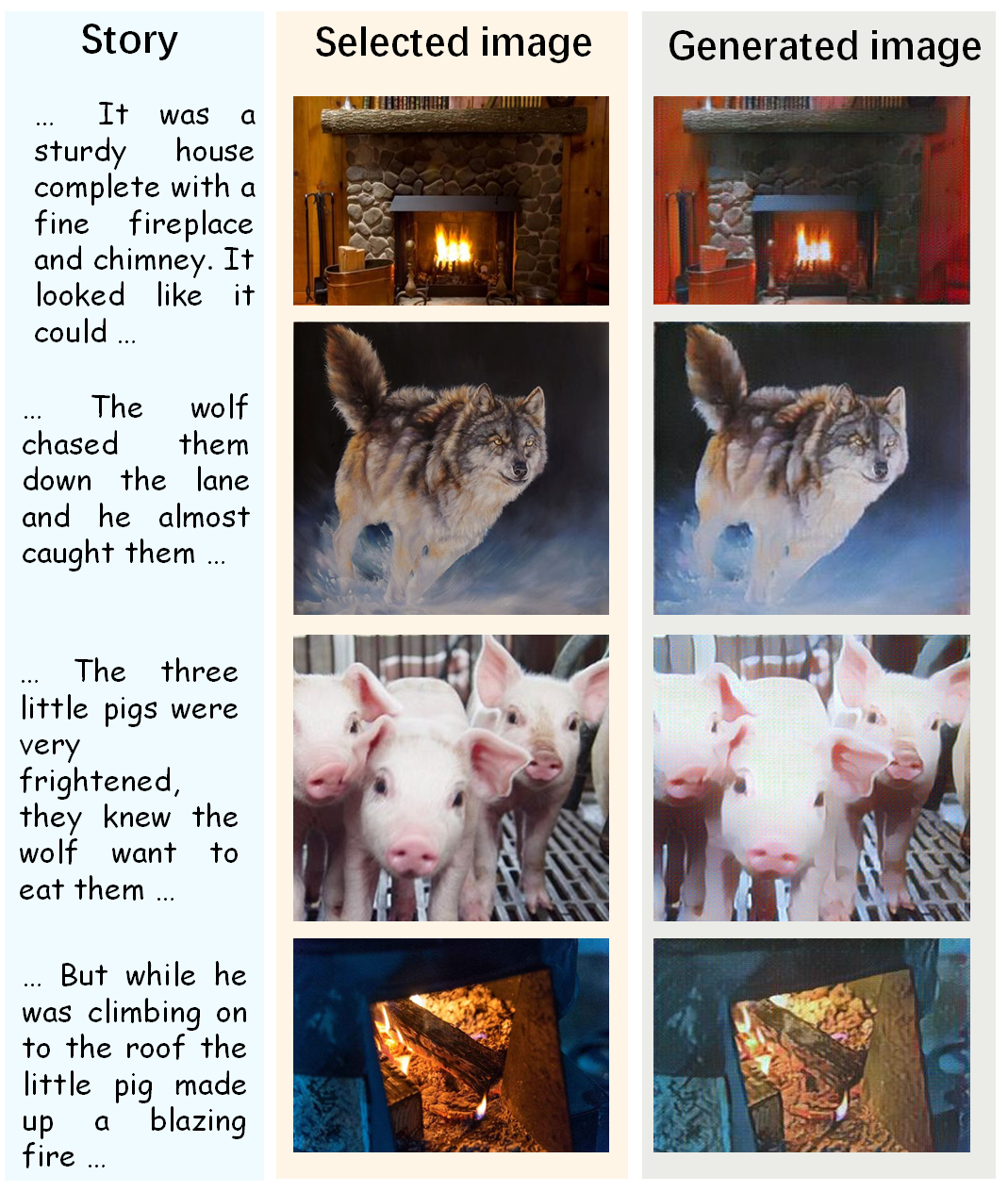}
\end{center}
   \caption{An example of story visualization.}
\label{fig:visual}
\end{figure}

We train our model on our coarse text-illustration dataset for 500 epochs. 
We select the original images from the search engine based on the story, and then regenerate the expected style.
An example of our work is shown in Fig~\ref{fig:visual}, where the classic fairy story "Three Pigs" is taken as input text, and the cartooning textures and colors are synthesized for the selected images.
Please see the supplementary materials for more examples.

\section{Conclusions}
\label{sec:con}
In this paper, we have proposed a novel text-based style generation task for coarsely matched texts.
We designed a generative adversarial network for text-based style generation (TSG-GAN), which exploited the image content feature and the text feature with a multi-modality style synthesis module.
A set of discriminators was further designed for stable training.
In ablation studies, we selected several indexes to demonstrate the effectiveness of our method in generating diverse and high-quality results.
Qualitative and quantitative comparisons showed our method outperformed state-of-the-art methods in generating consistent images for coarsely matched texts. 
Our framework has been successfully applied to text-image alignment and story visualization.

{\appendix
Here we show the training procedure, more results for ablation studies and experiments, and more details of user study.
}

\section{training procedure}
\begin{figure}[h]
\includegraphics[width=1\linewidth]{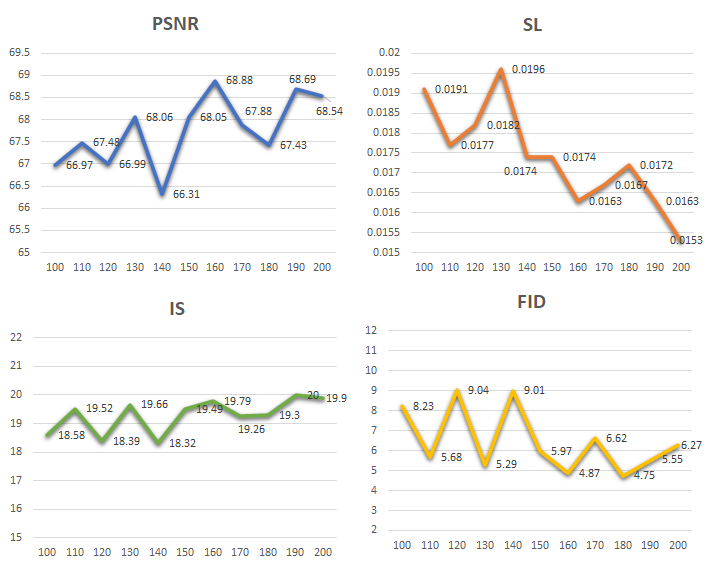}
\caption{The performance of the TSG-GAN during the training. Among the indexes, higher $IS$ and $PSNR$, and lower $SL$ and $FID$ mean better results.}
\label{fig:epoch}
\end{figure}

We summarize the TSG-GAN training procedure with Tab.~\ref{tab:algo}.
We calculate the performance of our model during the training (see Fig.~\ref{fig:epoch}).
We choose to train the model for 160 epochs, at which time it shares the highest PSNR with the relatively higher IS and the relatively lower SL and FID.

\begin{table}[H]
\centering
\caption{TSG-GAN Training Algorithm.}
\begin{tabular}{l}
\hline
\textbf{Input:} 
batch size $B$; \\
the hyper-parameter of the loss function $\lambda$; \\
the hyper-parameters of the Adam optimizer $\beta_1$, $\beta_2$; \\
generator, discriminator, text encoder parameters $\theta_G$, $\theta_D$, $\theta_T$; \\ 
generator, discriminator, text encoder learning rates $l_G$, $l_D$, $l_T$. \\
\hline
\textbf{for} the number of training steps \textbf{do} \\
\quad \quad \textbf{for} t = 1,2 \textbf{do}  \\
\quad \quad \quad \quad Sample $\{z_i\}_{i=1}^{B}\sim p_(z)$ \\
\quad \quad \quad \quad Sample $\{(I_i,h_i,e_i)\}_{i=1}^{B}\sim p_{data}(I,h,e)$ \\
\quad \quad \quad \quad $(L_D,L_S) \leftarrow \frac{1}{B}\sum^{B}_{i=1}(L_D(I_i,h_i,e_i), L_S(h_i))$ \\
\quad \quad \quad \quad $L_{D_{total}} \leftarrow L_D + \lambda L_S$ \\
\quad \quad \quad \quad $\theta_D \leftarrow Adam(L_{D_{total}},l_D,\beta_1,\beta_2)$ \\
\quad \quad \textbf{end for} \\
\quad \quad Sample $\{z_i\}_{i=1}^{B}\sim p(z)$ \\
\quad \quad Sample $\{(I_i,h_i,e_i)\}_{i=1}^{B}\sim p_{data}(I,h,e)$ \\
\quad \quad $(L_G,L_S) \leftarrow \frac{1}{B}\sum^{B}_{i=1}(L_G(I_i,h_i,e_i), L_S(h_i))$ \\
\quad \quad $L_{G_{total}} \leftarrow L_G + \lambda L_S$ \\
\quad \quad $\theta_G \leftarrow  Adam(L_{G_{total}},l_G,\beta_1,\beta_2)$ \\
\quad \quad $\theta_T \leftarrow Adam(L_{G_{total}},l_T,\beta_1,\beta_2)$ \\
\textbf{end for}\\
\hline
\end{tabular}
\label{tab:algo}
\end{table}

\section{Ablation studies}
We design additional ablation studies for the discriminator and the multi-modality style synthesis module. 
We also show a comparison of the results on different training epochs.

\subsection{Additional studies for the proposed multi-modality style synthesis model}

In the multi-modality style synthesis model, we reduce the spatial size of $v_0$ and get $v_{ca} \in \mathbb{R}^{C_1\times H_1\times W_1}, v_{sa} \in \mathbb{R}^{C_2\times H_2\times W_2}$. 
One reason for this option is to reduce the computational cost. 
Different from traditional image encoders in text-based image synthesis, Artflow~\cite{AnHuang-800} encodes the image feature to a higher spatial size ($64 \times 64$). 
For instance, in \cite{XuZhang-649}, the spatial size of the image feature is $17 \times 17$. 

\begin{figure*}
\includegraphics[width=1\linewidth]{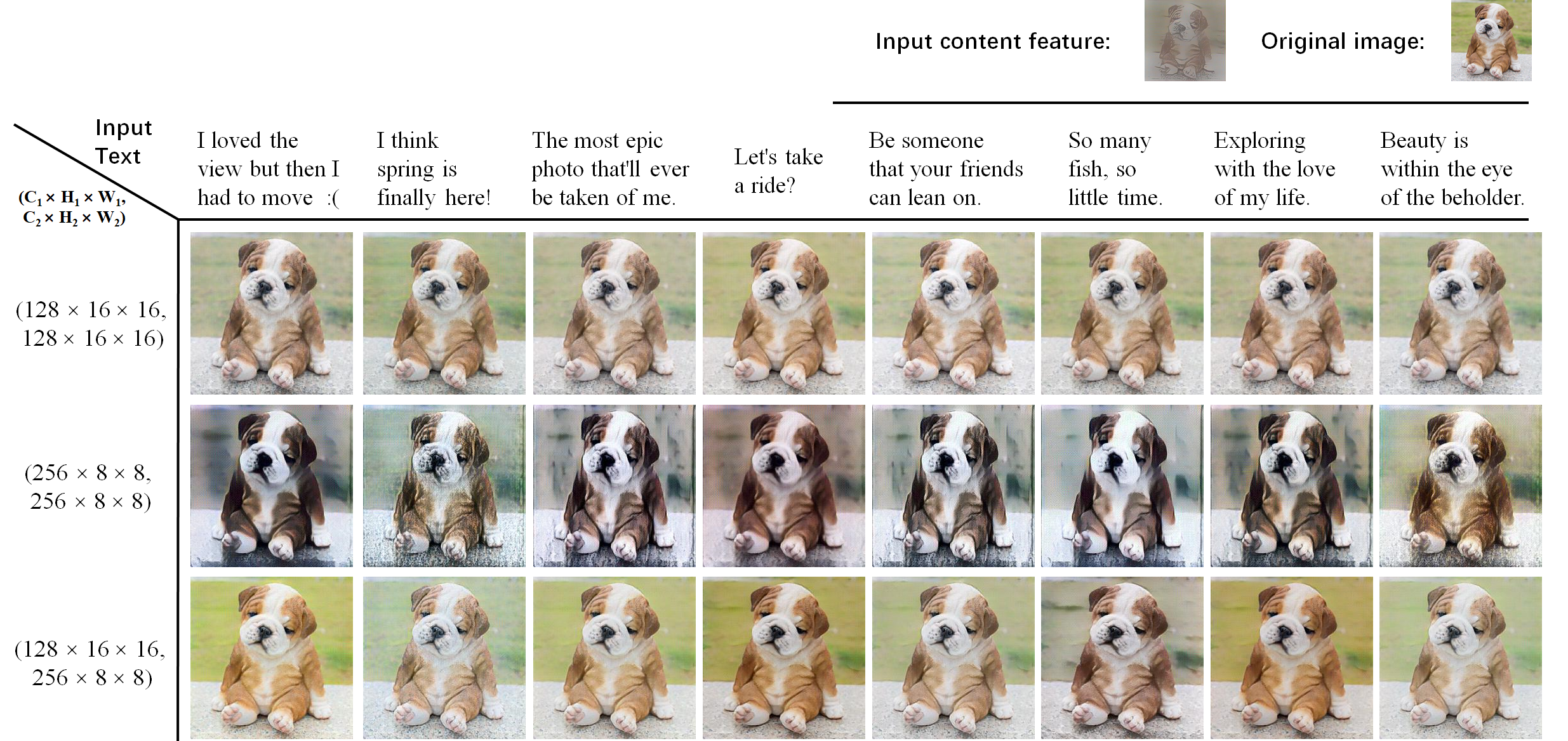}
\caption{Additional ablation study for the proposed multi-modality style synthesis model. We show the results under different values of $(C_1\times H_1\times W_1$, $C_2\times H_2\times W_2)$.}
\label{fig:com1}
\end{figure*}

We set $(C_1\times H_1\times W_1$, $C_2\times H_2\times W_2)$ to different values and train the model.
As shown in Tab.~\ref{tab:abla1}, the higher the spatial size, the better the performance. 
Note that we do not choose the model which has the best performance on the selected indexes, since there is a trade-off between the image quality and the style abundance.
More specifically, although the image quality of the first row is slightly better than the third row, it can only generate highly similar images under different input texts (see Fig.~\ref{fig:com1}). 
Although the second row shows a higher abundance under different input texts, their image quality is too low to be used.
Thus we set $(C_1\times H_1\times W_1$, $C_2\times H_2\times W_2)$ to $(128\times 16\times 16$, $256\times 8\times 8)$.

\begin{table}
\centering
\caption{Additional ablation studies for the proposed multi-modality style synthesis model: inception scores ($IS$), Fr$\acute{e}$chet Inception Distance ($FID$), Style Loss ($SL$), Peak Signal to Noise Ratio ($PSNR$) of our approaches on the Contextual Caption dataset.
}
\begin{tabular}{llrrrr}
\hline
$C_1\times H_1\times W_1$ & $C_2\times H_2\times W_2$ & IS & FID & SL & PSNR\\
\hline
$128 \times 16 \times 16$ & $ 128 \times 16 \times 16$ & 20.4 & 4.30 & 0.0131 & 70.38 \\
$256 \times 8 \times 8$ & $ 256 \times 8 \times 8$ & 13.6 & 24.3 & 0.0325 & 62.5 \\
$128 \times 16 \times 16$ & $ 256 \times 8 \times 8$ & \textbf{19.8} & \textbf{4.87} & \textbf{0.0163} & \textbf{68.9} \\
\hline
\end{tabular}
\label{tab:abla1}
\end{table}

\begin{table}
\centering
\caption{Additional studies for the loss function: inception scores ($IS$), Fr$\acute{e}$chet Inception Distance ($FID$), Style Loss ($SL$), Peak Signal to Noise Ratio ($PSNR$) of our approaches on the Contextual Caption dataset. Among the indexes, higher $IS$ and $PSNR$, and lower $SL$ and $FID$ mean better results.
}
\begin{tabular}{cccrrrr}
\hline
$D_s,D_{cs}$ & $L_s$ & $D_i,D_{ci}$ & IS & FID & SL & PSNR\\
\hline
\checkmark & & & 7.83 & 69.3 & 0.0462 & 59.21 \\
\checkmark & \checkmark & & 16.4 & 16.9 & 0.0171 & 64.6 \\
\checkmark & \checkmark & \checkmark & \textbf{19.8} & \textbf{4.87} & \textbf{0.0163} & \textbf{68.9} \\
\hline
\end{tabular}
\label{tab:abla2}
\end{table}

\begin{figure*}
\includegraphics[width=1\linewidth]{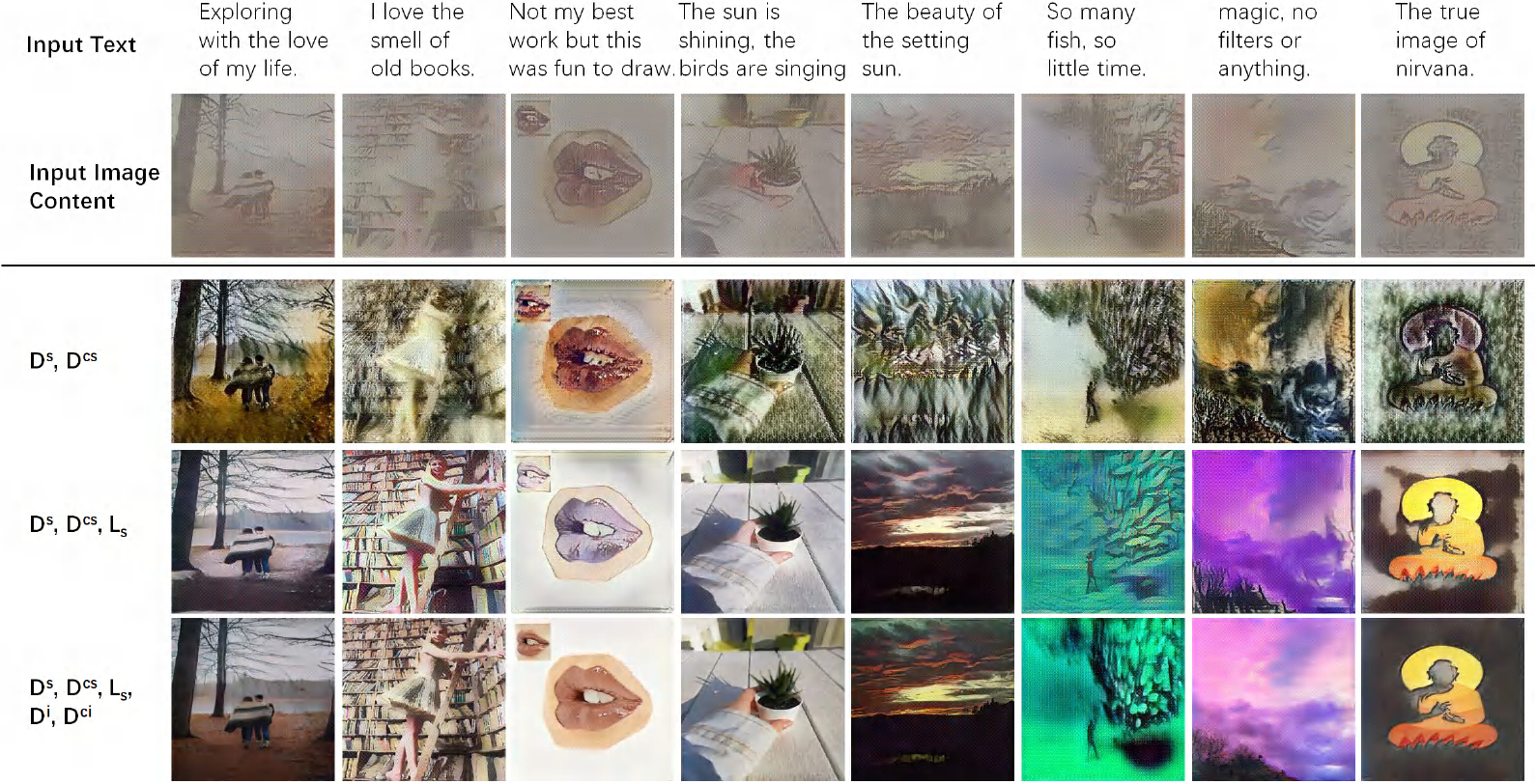}
\caption{Additional ablation study for the proposed loss function.}
\label{fig:com2}
\end{figure*}

\subsection{Effectiveness of the loss function}

We compare different settings of the loss function. 
As shown in Tab.~\ref{tab:abla2}, $L_s$ makes training more stable and gets better results.
$D_i$ and $D_{ci}$ further improve the image quality.
We also show examples of these settings in Fig.~\ref{fig:com2} to make a more intuitive comparison. 
%

\section{User study}

\begin{figure}
\includegraphics[width=1\linewidth]{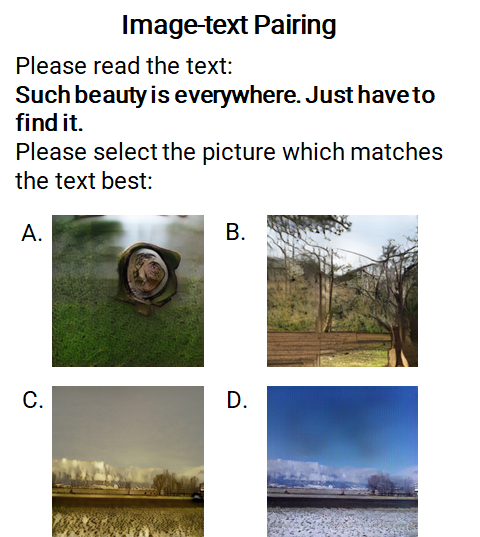}
\caption{The interface of user study A.}
\label{fig:in1}
\end{figure}

Our user study has two parts. 
We design part A to measure the degree of the correspondence between the original texts and the generated images.
The interface of the questionnaire in part A can be found in Fig.~\ref{fig:in1}.

\begin{figure}
\includegraphics[width=1\linewidth]{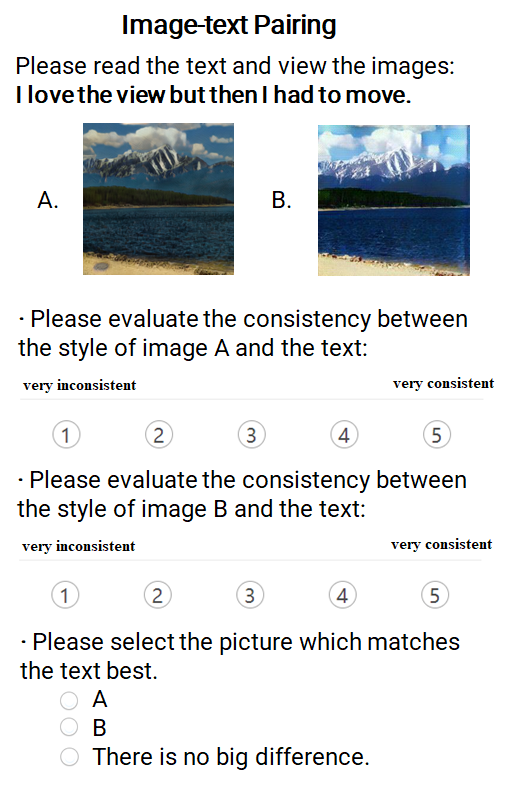}
\caption{The interface of user study B.}
\label{fig:in2}
\end{figure}

Part B is designed to measure the degree of consistency between a random input text and its generated image.
We also show the interface of questionnaire B in Fig.~\ref{fig:in2}.

\section{Additional results}
\subsection{Additional results of TSG-GAN}
\begin{figure*}
\includegraphics[width=1\linewidth]{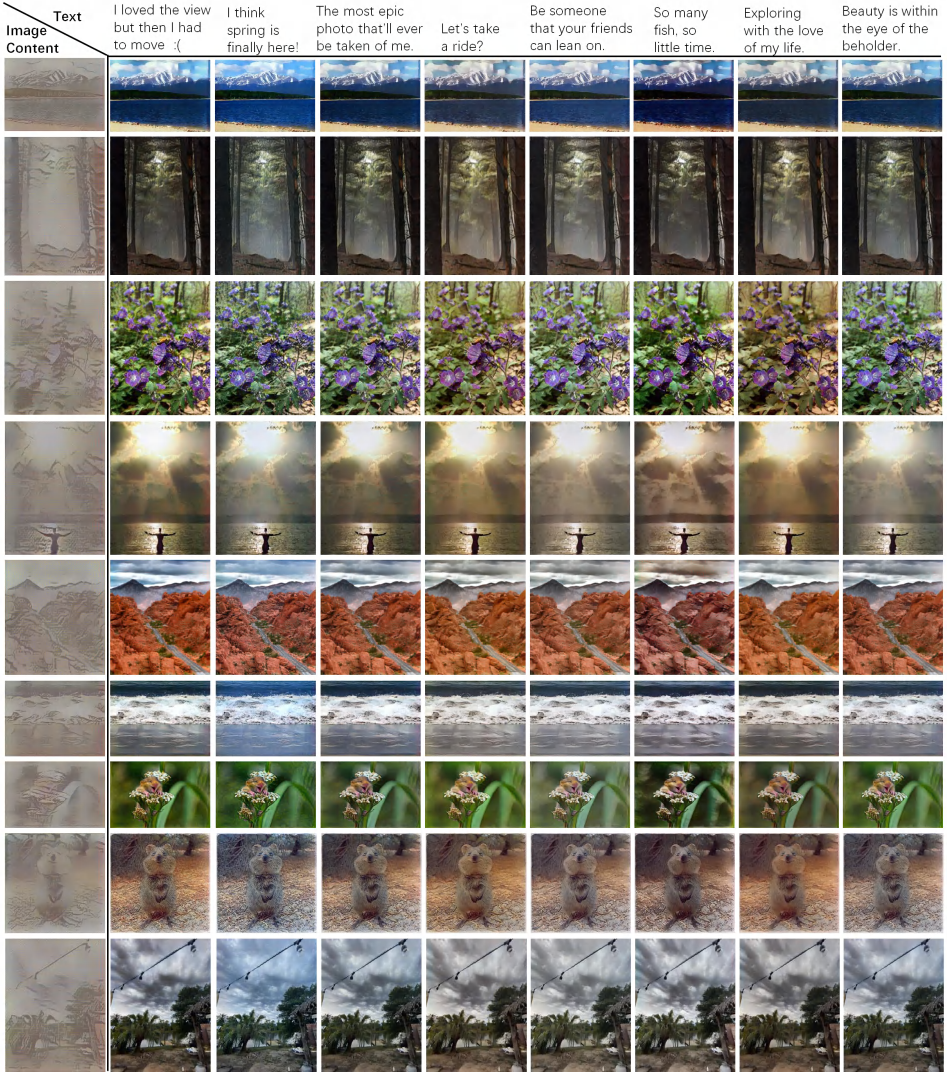}
\caption{Additional results of TSG-GAN.}
\label{fig:add}
\end{figure*}

We show additional results of our TSG-GAN on the Contextual Caption dataset~\cite{ChowdhuryBhowmik-758} in Fig.~\ref{fig:add}.

\begin{figure*}
\includegraphics[width=1\linewidth]{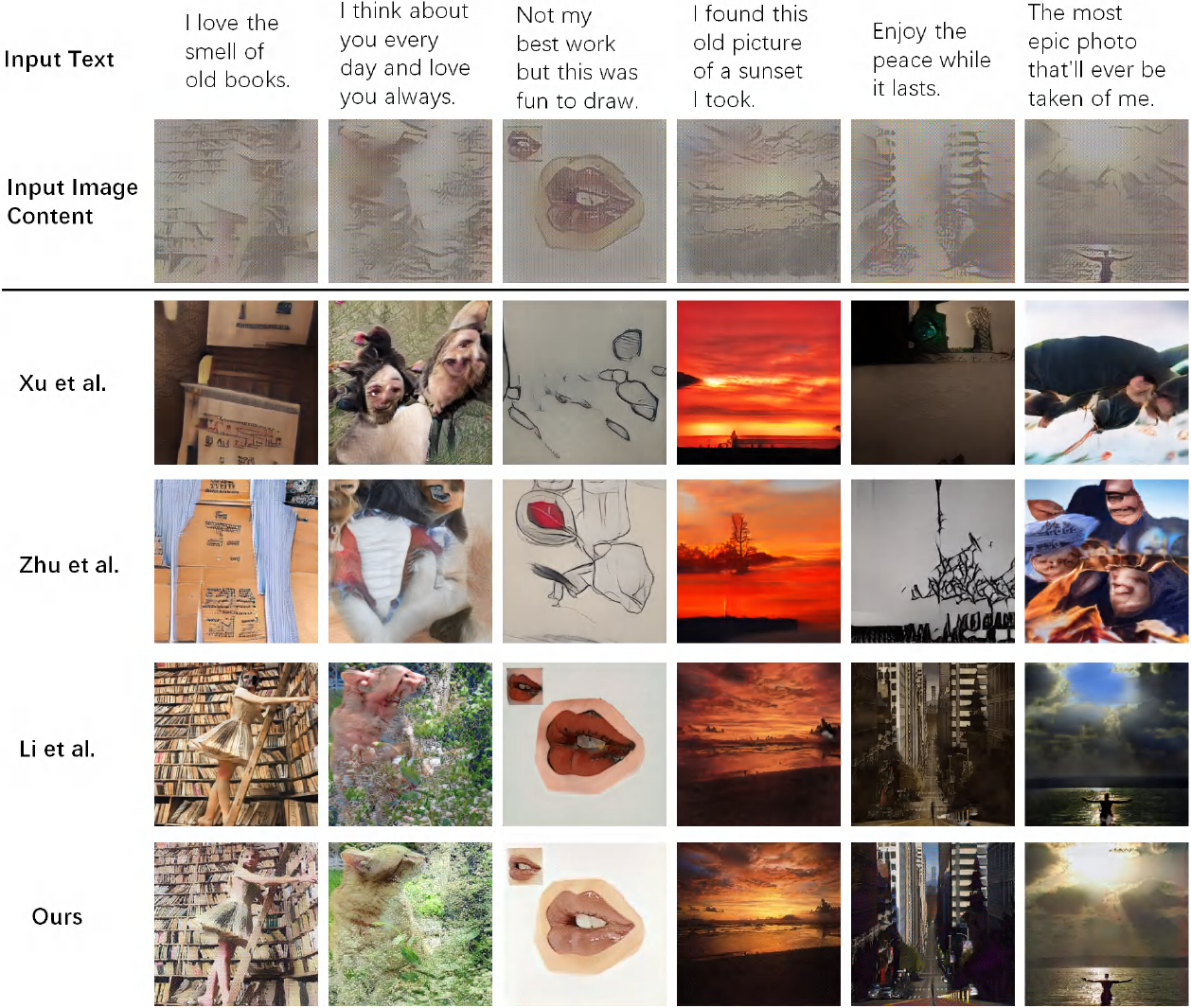}
\caption{Additional comparison results. The first row shows the result of \cite{XuZhang-649}, the second row shows the result of \cite{FengNiu-807}, the third row shows the result of \cite{LiQi-647}, and the last row shows the result of ours.}
\label{fig:add1}
\end{figure*}

\subsection{Additional comparison results}
We show additional comparison results of \cite{XuZhang-649}, \cite{FengNiu-807}, \cite{LiQi-647}, and ours on the Contextual Caption dataset in Fig.~\ref{fig:add1}.
As we can see in columns 1 and 4, when the input text includes an explicit object, the methods of \cite{XuZhang-649} and \cite{FengNiu-807} can generate meaningful results.

However, when there is no clear object mentioned (columns 2, 5, and 6), the results of \cite{XuZhang-649} and \cite{FengNiu-807} are distorted and less meaningful.
In comparison, \cite{LiQi-647} can generate meaningful images when there is no explicit object, but the texture of the results is not good (column 6).
Our method can generate high-quality images with fine textures.

\begin{figure*}
\includegraphics[width=1\linewidth]{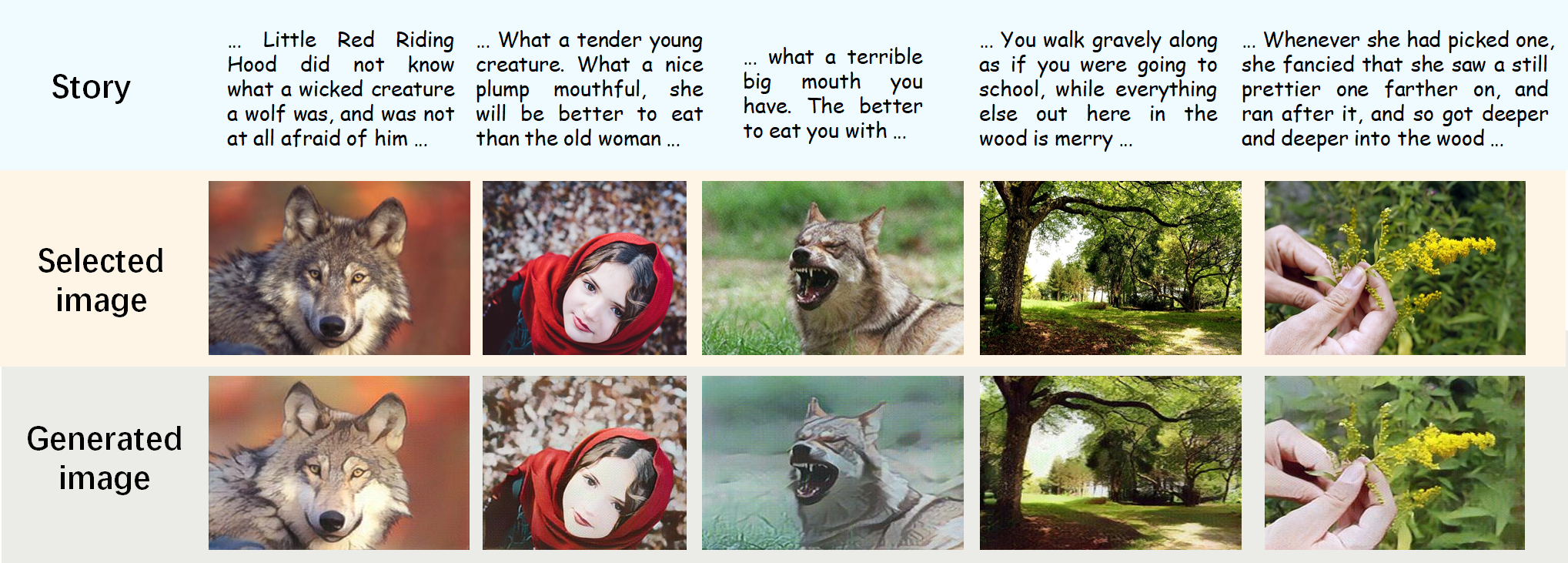}
\caption{An example of story visualization on "Little Red Riding Hood".}
\label{fig:sto2}
\end{figure*}

\subsection{Additional results for story visualization}
We train our model with a self-built dataset, which is collected from the illustration website~\footnote{\url{https://www.theispot.com/}}.
We chose two classic fairy stories for story visualization: "Three Pigs" and "Little Red Riding Hood".

In detail, we first select several images from the search engine.
Along with the input text, the images are then processed by the TSG-GAN.
More results can be found in Fig.~\ref{fig:sto2}.

\vfill
\bibliographystyle{IEEEtran}
\bibliography{bare_jrnl_new_sample4}

\begin{thebibliography}{10}
\providecommand{\url}[1]{#1}
\csname url@samestyle\endcsname
\providecommand{\newblock}{\relax}
\providecommand{\bibinfo}[2]{#2}
\providecommand{\BIBentrySTDinterwordspacing}{\spaceskip=0pt\relax}
\providecommand{\BIBentryALTinterwordstretchfactor}{4}
\providecommand{\BIBentryALTinterwordspacing}{\spaceskip=\fontdimen2\font plus
\BIBentryALTinterwordstretchfactor\fontdimen3\font minus \fontdimen4\font\relax}
\providecommand{\BIBforeignlanguage}[2]{{%
\expandafter\ifx\csname l@#1\endcsname\relax
\typeout{** WARNING: IEEEtran.bst: No hyphenation pattern has been}%
\typeout{** loaded for the language `#1'. Using the pattern for}%
\typeout{** the default language instead.}%
\else
\language=\csname l@#1\endcsname
\fi
#2}}
\providecommand{\BIBdecl}{\relax}
\BIBdecl

\bibitem{XuZhang-649}
T.~Xu, P.~Zhang, Q.~Huang, H.~Zhang, Z.~Gan, X.~Huang, and X.~He, ``Attngan: Fine-grained text to image generation with attentional generative adversarial networks,'' in \emph{Proc. CVPR}, 2018, pp. 1316--1324.

\bibitem{zhao2019image}
B.~Zhao, L.~Meng, W.~Yin, and L.~Sigal, ``Image generation from layout,'' in \emph{Proc. CVPR}, 2019, pp. 8584--8593.

\bibitem{johnson2018image}
J.~Johnson, A.~Gupta, and L.~Fei-Fei, ``Image generation from scene graphs,'' in \emph{Proc. CVPR}, 2018, pp. 1219--1228.

\bibitem{KohBaldridge-720}
J.~Y. Koh, J.~Baldridge, H.~Lee, and Y.~Yang, ``Text-to-image generation grounded by fine-grained user attention,'' in \emph{Proc. WACV}, 2021, pp. 237--246.

\bibitem{DongYu-654}
H.~Dong, S.~Yu, C.~Wu, and Y.~Guo, ``Semantic image synthesis via adversarial learning,'' in \emph{Proc. ICCV}, 2017, pp. 5706--5714.

\bibitem{LiQi-647}
B.~Li, X.~Qi, T.~Lukasiewicz, and P.~H. Torr, ``Manigan: Text-guided image manipulation,'' in \emph{Proc. CVPR}, 2020, pp. 7880--7889.

\bibitem{lin2014microsoft}
T.-Y. Lin, M.~Maire, S.~Belongie, J.~Hays, P.~Perona, D.~Ramanan, P.~Doll{\'a}r, and C.~L. Zitnick, ``Microsoft coco: Common objects in context,'' in \emph{Proc. ECCV}, 2014, pp. 740--755.

\bibitem{nilsback2008automated}
M.-E. Nilsback and A.~Zisserman, ``Automated flower classification over a large number of classes,'' in \emph{2008 Sixth Indian Conference on Computer Vision, Graphics \& Image Processing}.\hskip 1em plus 0.5em minus 0.4em\relax IEEE, 2008, pp. 722--729.

\bibitem{wah2011caltech}
C.~Wah, S.~Branson, P.~Welinder, P.~Perona, and S.~Belongie, ``The caltech-ucsd birds-200-2011 dataset,'' 2011.

\bibitem{ReedAkata-855}
S.~Reed, Z.~Akata, X.~Yan, L.~Logeswaran, B.~Schiele, and H.~Lee, ``Generative adversarial text to image synthesis,'' in \emph{Proc. ICML}, 2016, pp. 1060--1069.

\bibitem{FengNiu-807}
F.~Feng, T.~Niu, R.~Li, and X.~Wang, ``Modality disentangled discriminator for text-to-image synthesis,'' \emph{{IEEE} Trans. Multimedia}, vol.~24, pp. 2112--2124, 2021.

\bibitem{zhu2019dm}
M.~Zhu, P.~Pan, W.~Chen, and Y.~Yang, ``Dm-gan: Dynamic memory generative adversarial networks for text-to-image synthesis,'' in \emph{Proc. CVPR}, 2019, pp. 5802--5810.

\bibitem{LeeKim-768}
S.~Lee, D.~Kim, and B.~Han, ``Cosmo: Content-style modulation for image retrieval with text feedback,'' in \emph{Proc. CVPR}, 2021, pp. 802--812.

\bibitem{sutton2016color}
T.~M. Sutton and J.~Altarriba, ``Color associations to emotion and emotion-laden words: A collection of norms for stimulus construction and selection,'' \emph{Behavior research methods}, vol.~48, no.~2, pp. 686--728, 2016.

\bibitem{chen2020image}
T.~Chen, W.~Xiong, H.~Zheng, and J.~Luo, ``Image sentiment transfer,'' in \emph{Proc. ACM MM}, 2020, pp. 4407--4415.

\bibitem{ChowdhuryBhowmik-758}
S.~N. Chowdhury, R.~Bhowmik, H.~Ravi, G.~de~Melo, S.~Razniewski, and G.~Weikum, ``Exploiting image--text synergy for contextual image captioning,'' in \emph{Proc. LANTERN}, 2021, pp. 30--37.

\bibitem{chowdhury2021sandi}
S.~N. Chowdhury, S.~Razniewski, and G.~Weikum, ``Sandi: Story-and-images alignment,'' in \emph{Proc. ACL}, 2021, pp. 989--999.

\bibitem{LiGan-661}
Y.~Li, Z.~Gan, Y.~Shen, J.~Liu, Y.~Cheng, Y.~Wu, L.~Carin, D.~Carlson, and J.~Gao, ``Storygan: A sequential conditional gan for story visualization,'' in \emph{Proc. CVPR}, 2019, pp. 6329--6338.

\bibitem{Song-663}
Y.-Z. Song, Z.~R. Tam, H.-J. Chen, H.-H. Lu, and H.-H. Shuai, ``Character-preserving coherent story visualization,'' in \emph{Proc. ECCV}, 2020, pp. 18--33.

\bibitem{ZhangXu-851}
H.~Zhang, T.~Xu, H.~Li, S.~Zhang, X.~Wang, X.~Huang, and D.~N. Metaxas, ``Stackgan: Text to photo-realistic image synthesis with stacked generative adversarial networks,'' in \emph{Proc. ICCV}, 2017, pp. 5907--5915.

\bibitem{ZhangKoh-806}
H.~Zhang, J.~Y. Koh, J.~Baldridge, H.~Lee, and Y.~Yang, ``Cross-modal contrastive learning for text-to-image generation,'' in \emph{Proc. CVPR}, 2021, pp. 833--842.

\bibitem{NamKim-655}
S.~Nam, Y.~Kim, and S.~J. Kim, ``Text-adaptive generative adversarial networks: manipulating images with natural language,'' in \emph{Proc. NIPS}, 2018, pp. 42--51.

\bibitem{PatashnikWu-686}
O.~Patashnik, Z.~Wu, E.~Shechtman, D.~Cohen-Or, and D.~Lischinski, ``Styleclip: Text-driven manipulation of stylegan imagery,'' in \emph{Proc. ICCV}, 2021, pp. 2085--2094.

\bibitem{radford2021learning}
A.~Radford, J.~W. Kim, C.~Hallacy, A.~Ramesh, G.~Goh, S.~Agarwal, G.~Sastry, A.~Askell, P.~Mishkin, J.~Clark \emph{et~al.}, ``Learning transferable visual models from natural language supervision,'' \emph{arXiv preprint arXiv:2103.00020}, 2021.

\bibitem{KarrasLaine-583}
T.~Karras, S.~Laine, and T.~Aila, ``A style-based generator architecture for generative adversarial networks,'' in \emph{Proc. CVPR}, 2019, pp. 4401--4410.

\bibitem{dhamo2020semantic}
H.~Dhamo, A.~Farshad, I.~Laina, N.~Navab, G.~D. Hager, F.~Tombari, and C.~Rupprecht, ``Semantic image manipulation using scene graphs,'' in \emph{Proc. CVPR}, 2020, pp. 5213--5222.

\bibitem{huang2017arbitrary}
X.~Huang and S.~Belongie, ``Arbitrary style transfer in real-time with adaptive instance normalization,'' in \emph{Proc. ICCV}, 2017, pp. 1501--1510.

\bibitem{liu2020describe}
Y.~Liu, M.~De~Nadai, D.~Cai, H.~Li, X.~Alameda-Pineda, N.~Sebe, and B.~Lepri, ``Describe what to change: A text-guided unsupervised image-to-image translation approach,'' in \emph{Proc. ACM MM}, 2020, pp. 1357--1365.

\bibitem{li2017universal}
Y.~Li, C.~Fang, J.~Yang, Z.~Wang, X.~Lu, and M.-H. Yang, ``Universal style transfer via feature transforms,'' \emph{Advances in neural information processing systems}, vol.~30, 2017.

\bibitem{JingLiu-726}
Y.~Jing, X.~Liu, Y.~Ding, X.~Wang, E.~Ding, M.~Song, and S.~Wen, ``Dynamic instance normalization for arbitrary style transfer,'' in \emph{Proc. AAAI}, vol.~34, no.~04, 2020, pp. 4369--4376.

\bibitem{yang2020fda}
Y.~Yang and S.~Soatto, ``Fda: Fourier domain adaptation for semantic segmentation,'' in \emph{Proc. CVPR}, 2020, pp. 4085--4095.

\bibitem{xu2021fourier}
Q.~Xu, R.~Zhang, Y.~Zhang, Y.~Wang, and Q.~Tian, ``A fourier-based framework for domain generalization,'' in \emph{Proc. CVPR}, 2021, pp. 14\,383--14\,392.

\bibitem{jing2020dynamic}
Y.~Jing, X.~Liu, Y.~Ding, X.~Wang, E.~Ding, M.~Song, and S.~Wen, ``Dynamic instance normalization for arbitrary style transfer,'' in \emph{Proc. AAAI}, vol.~34, no.~04, 2020, pp. 4369--4376.

\bibitem{sheng2018avatar}
L.~Sheng, Z.~Lin, J.~Shao, and X.~Wang, ``Avatar-net: Multi-scale zero-shot style transfer by feature decoration,'' in \emph{Proc. CVPR}, 2018, pp. 8242--8250.

\bibitem{svoboda2020two}
J.~Svoboda, A.~Anoosheh, C.~Osendorfer, and J.~Masci, ``Two-stage peer-regularized feature recombination for arbitrary image style transfer,'' in \emph{Proc. CVPR}, 2020, pp. 13\,816--13\,825.

\bibitem{AnHuang-800}
J.~An, S.~Huang, Y.~Song, D.~Dou, W.~Liu, and J.~Luo, ``Artflow: Unbiased image style transfer via reversible neural flows,'' in \emph{Proc. CVPR}, 2021, pp. 862--871.

\bibitem{kingma2014adam}
D.~P. Kingma and J.~Ba, ``Adam: A method for stochastic optimization,'' \emph{arXiv preprint arXiv:1412.6980}, 2014.

\bibitem{LiuLebret-679}
F.~Liu, R.~Lebret, D.~Orel, P.~Sordet, and K.~Aberer, ``Upgrading the newsroom: An automated image selection system for news articles,'' \emph{{ACM} Trans. Multimedia.}, vol.~16, no.~3, pp. 1--28, 2020.

\bibitem{ZhangWang-746}
Z.~Zhang, J.~Wang, A.~Jatowt, Z.~Sun, S.-P. Lu, and Z.~Yang, ``News content completion with location-aware image selection,'' in \emph{Proc. AAAI}, vol.~35, no.~16, 2021, pp. 14\,498--14\,505.

\end{thebibliography}
\end{document}